\documentclass[11pt]{article}
\usepackage{hyperref}

\usepackage[final]{acl}

\usepackage{times}
\usepackage{latexsym}
\usepackage{tikz}
\usepackage{pgfplots}
\pgfplotsset{compat=1.18}
\usepackage[table,dvipsnames]{xcolor}
\usepackage{multirow}
\usepackage{enumitem}

\usepackage[T1]{fontenc}

\usepackage[utf8]{inputenc}

\usepackage{microtype}

\usepackage{inconsolata}

\usepackage{graphicx}
\usepackage{pifont}
\usepackage{tcolorbox}
\usepackage{algorithm}
\usepackage{algcompatible}
\usepackage{algorithmicx}
\usepackage{algpseudocode}
\usepackage{amsmath}
\usepackage{booktabs}
\usepackage{enumitem}
\usepackage{amssymb}
\usepackage{pifont}

\usepackage{tikz}
\usetikzlibrary{arrows.meta, positioning, shapes.geometric}

\usepackage{todonotes}

\title{Explanation Quality Assessment as Ranking with Listwise Rewards}

\author{%
  Thomas Bailleux\textsuperscript{1}\thanks{These authors contributed equally and are co-first authors.},
  Tanmoy Mukherjee\textsuperscript{1}\footnotemark[1], {\bf Emmanuel Lonca}\textsuperscript{1},\\
  {\bf Pierre Marquis}\textsuperscript{1,2},
  {\bf Zied Bouraoui}\textsuperscript{1}
  \\[4pt]
  \textsuperscript{1} CRIL, Univ. Artois \& CNRS, France, \quad
  \textsuperscript{2} Institut Universitaire de France
  \\
  \texttt{\{mukherjee,bailleux,marquis,bouraoui\}@cril.fr}
}

\begin{document}
\maketitle
\begin{abstract}
We reformulate explanation quality assessment as a ranking problem rather than a generation problem. Instead of optimizing models to produce a single “best” explanation token-by-token, we train reward models to discriminate among multiple candidate explanations and learn their relative quality. Concretely, we construct per-instance candidate sets with graded quality levels and train listwise and pairwise ranking models (ListNet, LambdaRank, RankNet) to preserve ordinal structure and avoid score compression typical of pointwise regression or binary preference objectives. We observe three findings: First, ranking losses consistently outperform regression on score separation across all domains tested. Second, the optimal ranking loss depends on data characteristics: listwise objectives excel with well-separated quality tiers, while pairwise methods are more robust to noisy natural annotations. Third, when trained on carefully curated and well-structured data, small encoder models can match models that are orders of magnitude larger, suggesting that data quality matters more than model scale. Finally, when used as rewards in policy optimization, ranking-based scores enable stable convergence in settings where regression-based rewards fail entirely. Code and data are available at: \url{https://github.com/Tankiit/PPO_Learning_to_rank}
\end{abstract}

\section{Introduction}
\label{introduction}
Although Large Language Models (LLMs) can produce many plausible explanations for a given query, a central challenge lies in determining which explanations are actually better than others. Explanation quality is inherently graded: two explanations may both appear plausible while differing substantially in logical consistency, factual correctness, and relevance to the underlying query. Yet most existing approaches treat explanation evaluation either as binary classification or as a pointwise generation problem, thereby collapsing fine-grained quality distinctions. In this work, we instead formulate explanation quality assessment as a \emph{learning-to-rank} problem, in which models learn to order multiple candidate explanations by relative quality.

A key difficulty is that standard reward objectives do not preserve these graded distinctions well enough for policy optimization. In particular, regression and binary preference objectives tend to compress explanation scores into a narrow range, making high- and low-quality explanations insufficiently distinguishable. This weakens the reward signal used by Proximal Policy Optimization (PPO): when score separation is too small, advantage estimates become dominated by noise and policy updates become ineffective. By contrast, ranking objectives preserve ordinal structure and maintain substantially larger score separation, yielding rewards that better support policy learning.

Neural ranking models are well established in information retrieval~\cite{Nogueira2019MultiStageDR,zhuang2023rankt5}. Our contribution is therefore not ranking per se, but identifying why ranking matters for explanation quality assessment and policy optimization. Specifically, we show that score compression is a key failure mode of regression and preference-based reward modeling; that the most effective ranking objective depends on the structure and noise profile of the data; and that ranking objectives require quality-diverse candidate sets in order to preserve useful score separation.

We integrate ranking-based reward models into PPO so that generated explanations are evaluated using relative quality signals rather than compressed point estimates. We validate the approach on 50,000 human-annotated explanations from e-SNLI, whose gold-level annotations were collected with 87\% inter-annotator agreement and Fleiss' $\kappa = 0.72$ as 
reported by the e-SNLI authors~\cite{camburu2018esnli}. Our experiments span seven architectures, ranging from 110M to 7B parameters, and four NLI benchmarks. We focus on NLI because it provides rich explanation annotations and controlled validation settings, and we additionally test transfer to commonsense reasoning tasks to assess generalization beyond the primary domain. Figure~\ref{fig:pipeline-comparison} (Appendix) contrasts the generation-centric and ranking-based pipelines.
\begin{enumerate}[label=(\roman*)]
    \item We identify score compression as a failure mode in explanation reward modeling. Regression and binary preference objectives compress graded quality scores into a narrow range, which weakens PPO learning signal.
    
    \item We establish an end-to-end link between ranking objectives and policy optimization. We show that better preservation of score separation leads to more informative advantage estimates, faster PPO convergence, and higher explanation quality.

    \item We show that the effectiveness of ranking depends on both the loss and the data. Listwise and pairwise objectives behave differently depending on the structure and noise level of the candidate sets, and ranking alone is insufficient without quality-diverse training data.

    \item We introduce a graded dataset construction paradigm for explanation ranking. We recast NLI benchmarks as ranking tasks by building five-level 
quality hierarchies with overlapping score ranges, anchored to e-SNLI's human-authored gold explanations.
\end{enumerate}

\section{Related Work}
\label{sec:related-work}

Our work connects to research on explanation evaluation, human judgment modeling, and ranking-based reinforcement learning from feedback. 

\noindent\textbf{Explanation Evaluation.}
Recent work has advanced the generation and evaluation of explanations through structured and abductive reasoning.   \textit{DecompRC}~\cite{min2019multihopreadingcomprehensionquestion} and \textit{Least-to-Most Prompting}~\cite{zhou2023leasttomostpromptingenablescomplex} improve interpretability by decomposing reasoning into sub-steps, while \textit{DRFACT}~\cite{lin2021differentiableopenendedcommonsensereasoning} and \textit{GEAR}~\cite{he2025geargeneralevaluationframework} formalize open-ended and abductive reasoning using criteria such as consistency and diversity. We integrates quality supervision directly into training 
via ranking objectives, quantifying quality for continuous optimization.

\noindent\textbf{Human Judgment and Disagreement.} 
\textit{ChaosNLI}~\cite{nie2020chaosnli} showed inherent annotation uncertainty. Following findings that ambiguity improves calibration ~\cite{uma2021learning,plank2022problem}, we induce controlled variance through overlapping 
score ranges, emulating natural disagreement without costly annotation.

\noindent\textbf{Policy Optimization with Learned Rewards.} Reward learning from preferences \citep{christiano2023deepreinforcementlearninghuman,ouyang2022traininglanguagemodelsfollow} aligns models  with human intent. We replace discrete ratings with continuous listwise supervision from graded rankings.

\noindent\textbf{Ranking and Preference Learning.}
We adapt ListNet, RankNet, LambdaRank, and ApproxNDCG that directly optimize for correct ordering rather than pointwise errors. Unlike text generation applications \cite{zhuang2023rankt5}, we target explanation  evaluation where quality exists on a continuum requiring specialized ranking objectives.


\section{Methodology}
\label{sec:methodology}
We start from the insight that explanation quality is inherently a \emph{ranking} problem rather than a regression or classification task. Instead, we rank existing explanations by quality.  

\noindent\textbf{Motivation.}
Current explanation datasets (e.g., e-SNLI, ChaosNLI, or $\delta$-NLI) reveal quality hierarchies, yet models reduce these to binary or
regressed values. Reward models trained with MSE compress rich scales into narrow ranges, producing negligible variance for PPO. This discards ranking-style supervision, limiting effective policy learning.

\noindent\textbf{From Generation to Ranking.}
Given an input $(p,h,y)$, where $p$ denotes the premise, $h$ the hypothesis, and $y \in \{\textit{entail}, \textit{contradict}, \textit{neutral}\}$, a model generates an explanation $e=(t_1,\ldots,t_n)$ token by token using $\pi_\theta(\cdot \,|\, t_{<i}, p, h, y)$, receives a scalar reward $R(e)$ upon completion, and updates parameters with PPO.  
This generation-centric setup has three problems: (i) sparse terminal rewards, (ii) binary or regressed quality signals, and (iii) a mismatch between generation and evaluation objectives.  
We instead use a ranking-based framework:  
for a question $q$ with candidate explanations $\mathcal{E}=\{e_1,\ldots,e_n\}$, we train a scoring function $f_\theta(q,e)\!\rightarrow\!\mathbb{R}$ to induce the correct quality ordering.  
Unlike pointwise regression, ranking losses directly optimize \emph{relative orderings}, providing dense, feedback-aligned supervision.

\noindent\textbf{Explanation quality.}
We define explanation quality as the extent to which a candidate explanation (i) is \emph{coherent} with the logical relation between the premise and hypothesis, (ii) is \emph{factually accurate} in its justification, and (iii) is \emph{relevant}—it directly answers “why does $y$ hold between $p$ and $h$?”. For the gold tier, this definition is operationalized through the 
e-SNLI human annotations; the four synthetic tiers operationalize 
it through deterministic templates. We do not assume human 
annotations are an objective gold standard, only that they are 
the appropriate reference audience for the gold tier.

\noindent\textbf{Ranking-Aware PPO.}
Within PPO, the ranking model serves as a reward estimator.  The normalized advantage is computed as
$\hat{A}_t = f_\theta(q_t, e_t) - V_\phi(q_t)$, 
where $V_\phi$ is the value baseline.  
Because $f_\theta$ captures comparative rather than absolute quality, $\hat{A}_t$ measures how much better a given explanation is relative to its peers, providing stable and interpretable gradients. This reframes explanation modeling as \emph{selection with structured feedback} rather than generation with sparse scalar rewards.

\noindent\textbf{Learning-to-Rank Integration.}
We integrate ranking via
(i)~\emph{Ranking Loss Functions:} ListNet, RankNet, and LambdaRank objectives that maintain score separation   
(ii)~\emph{Listwise Training:} on full explanation sets per query, using the complete graded signal instead of binary pairs;
(iii)~\emph{Ranking-Based Rewards}: PPO reward $r_t=f_\theta(q_t,e_t)$ preserves ordinal gaps.

\begin{table*}[t]
\centering
\scriptsize
\begin{tabular}{@{}llcccccccc@{}}
\toprule
\textbf{Cat.} & \textbf{Model} & \textbf{Params}
  & \textbf{NDCG@5} & \textbf{MAP} & \textbf{$\rho$}
  & \textbf{it/s} & \textbf{Time}
  & \textbf{Seq.} & \textbf{GFLOPs} \\
\midrule
\multirow{3}{*}{Enc.}
& BERT-base    & 110M & 0.873 & 0.827 & 0.724 & 18.2 & 2.7h & 512 & 22 \\
& \textbf{RoBERTa} & \textbf{125M}
               & \textbf{0.996} & \textbf{1.000} & \textbf{0.920}
               & \textbf{16.5} & \textbf{2.9h} & \textbf{512} & \textbf{25} \\
& DeBERTa-v3   & 184M & 0.912 & 0.827 & 0.798 & 15.3 & 3.2h & 512 & 37 \\
\midrule
\multirow{4}{*}{\shortstack{Dec.\\(4-bit)}}
& Phi-2        & 2.7B & 1.000 & 1.000 & 1.000 &  6.8 &  8.1h & 256 & 346 \\
& Falcon-7B    &   7B & 1.000 & 1.000 & 1.000 &  5.3 & 10.4h & 256 & 896 \\
& MPT-7B       &   7B & 1.000 & 1.000 & 1.000 &  5.5 & 10.1h & 256 & 896 \\
& Mistral-7B   &   7B & 1.000 & 1.000 & 1.000 &  5.1 & 10.8h & 256 & 896 \\
\bottomrule
\end{tabular}
\caption{Model comparison with ListNet loss. Data quality enables near-perfect discrimination regardless of scale. Seq.\ = effective
input length (tokens); GFLOPs = forward-pass compute per query$^\dagger$. Encoders process $2\times$ longer sequences at ${\sim}20\times$ lower compute than 7B decoders.}
\label{tab:models}
\end{table*}

\section{Experiments}
\label{sec:experiments}
We validate through experiments on: (1) reward modeling failure modes, (2) ranking vs. generation, (3) architecture selection, (4) data creation, and (5) PPO improvement.

\noindent\textbf{Task.}  
Given a query $q = (p, h, y)$ and candidate explanations $\mathcal{E} = \{e_1, \ldots, e_k\}$ with quality scores $s_i \in [0,1]$, the goal is to learn a scoring function $f_\theta(q, e) \!\to\! \mathbb{R}$ that orders explanations by quality.

\noindent\textbf{Models.}  
We compare \textit{Encoders:} BERT-base, RoBERTa-base, and DeBERTa-v3. \textit{Decoders} (4-bit quantized~\cite{dettmers2023qlora}): Phi-2, Falcon-7B, MPT-7B, and Mistral-7B. All use a pretrained backbone with a two-layer projection head (dropout 0.1).

\noindent\textbf{Objectives and Evaluation.}  
We compare MSE regression, binary classification (Bradley-Terry preference model \cite{rafailov2023dpo}), and
ranking losses: ListNet~\cite{cao2007listnet},
RankNet~\cite{burges2005ranknet}, and
ApproxNDCG~\cite{bruch2019approxndcg}. We evaluate with NDCG@$k$~\cite{jarvelin2002ndcg}, MAP, Spearman's~$\rho$, Kendall's~$\tau$, and score separation $\sigma(\hat{s})/\sigma(s)$.
Details are in App.~\ref{app:ranking loss}.

\noindent\textbf{Training.}  
Models are trained for 50 epochs with early stopping (patience~10) using AdamW and a 500-step warmup.   Encoders: batch~16, lr~2e-5; Decoders: batch~4, lr~1e-5. Further information about architecture is provided in App~\ref{app:model architecture}

\noindent\textbf{Computational fairness.}
Encoder and decoder models differ in tokeniser vocabularies and effective context windows: encoders process up to 512 tokens (bidirectional
attention), while decoder models are capped at 256 tokens (causal attention) to fit within 32\,GB GPU memory during 4-bit inference.
Table~\ref{tab:models} presents throughput (it/s), training time, and GFLOPs per forward pass to facilitate fair cross-architecture comparison; encoders attain substantially lower compute per pass despite processing much longer sequences.

\noindent\textbf{Datasets.} Our main evaluation benchmark is e-SNLI~\cite{camburu2018esnli}, which provides 50k human-written explanations and constitutes our primary pool of \textit{gold-tier} candidates. We additionally use MultiNLI (700 validation instances), Delta-NLI (68k), and WinoWhy (14k) to test generalization across NLI scenarios. To examine transfer outside the NLI domain, we further evaluate models trained exclusively on e-SNLI on three held-out commonsense benchmarks with zero additional fine-tuning: WinoGrande, CommonsenseQA, and StrategyQA (NDCG@5 scores in App.~\ref{app:analysis}).
 
\noindent\textbf{Graded Dataset Construction.}
Standard NLI datasets have zero score variance: all explanations for a given $(p, h)$ share one label, making ranking metrics degenerate. We convert them into ranking benchmarks by creating five quality-graded explanations per instance: \textit{gold}, \textit{good}, \textit{fair}, \textit{poor}, and \textit{nonsense}. \emph{Gold} explanations are human-authored e-SNLI annotations~\cite{camburu2018esnli} with 87\% inter-annotator agreement and Fleiss'~$\kappa = 0.72$. The other four levels use deterministic, label-aware templates (Table~\ref{tab:templates} in App.~\ref{app:graded_dataset}) and a content-aware heuristic scorer (Algorithm~\ref{alg:scoring}); their ordering is fixed by design, not by human judgment. Overlapping score ranges ensure adjacent levels partially intersect, adding realistic ambiguity. The resulting data is challenging yet structured, and the method applies to e-SNLI, $\delta$-NLI, and WinoWhy. Full construction details, score ranges, and validation metrics are in App.~\ref{app:graded_dataset}.

\noindent\textbf{Human Validation.} To illustrate how our graded gold ordering acts as a validation signal, consider Table~\ref{tab:qualitative} (premise: \textit{``A man is playing guitar on stage''}; hypothesis: \textit{``A musician is performing''}). The five candidates range from a premise-grounded gold explanation to a semantically irrelevant distractor. A ListNet-trained model recovers the gold ordering, whereas an MSE-trained model collapses them into a narrow score band and loses the ranking. On held-out e-SNLI queries, the pattern persists: ListNet preserves the gold ordering, RankNet partially recovers it, and MSE is only slightly above chance. Gold-tier candidates inherit their reliability from e-SNLI human annotations~\cite{camburu2018esnli}; tier scores and per-model agreement metrics are detailed in App.~\ref{app:analysis}.

\subsection{Results}

\begin{table}[t]
\centering
\scriptsize
\begin{tabular}{@{}lp{4cm}cc@{}}
\toprule
\textbf{Quality} & \textbf{Explanation (truncated)}
  & \textbf{MSE} & \textbf{ListNet} \\
\midrule
Gold
  & Playing guitar on stage is a form of musical performance.
  & 0.51 & 0.91 \\
Good
  & The premise entails the hypothesis: playing guitar = performing music.
  & 0.51 & 0.71 \\
Fair
  & The premise supports the hypothesis.
  & 0.50 & 0.52 \\
Poor
  & Contradiction: man $\neq$ musician. \textit{[wrong label]}
  & 0.49 & 0.32 \\
Nonsense
  & Penguin migration patterns suggest umbrella distribution.
  & 0.48 & 0.13 \\
\midrule
\multicolumn{2}{l}{\textit{Score range}}
  & \textbf{0.03} & \textbf{0.78} \\
\bottomrule
\end{tabular}
\caption{Score compression example (premise: \textit{``A man is playing guitar on stage''}; hypothesis: \textit{``A musician is
performing''}; label: entailment). 
}
\label{tab:qualitative}
\end{table}

\begin{table}[t]
\centering
\scriptsize
\begin{tabular}{@{}l@{}cccc@{}}
\toprule
\textbf{Loss} & \textbf{NDCG@5} & \textbf{Spearman $\rho$} & \textbf{Sep. Ratio} & \textbf{ Range} \\
\midrule
MSE Regression & 0.789 & 0.512 & 0.089 & 0.04 \\
Binary classification & 0.815 & 0.547 & 0.124 & 0.16 \\
RankNet & 0.856 & 0.682 & 0.341 & 0.43 \\
ApproxNDCG & 0.874 & 0.721 & 0.523 & 0.61 \\
\textbf{ListNet} & \textbf{0.996} & \textbf{0.920} & \textbf{0.920} & \textbf{0.85} \\
\bottomrule
\end{tabular}
\caption{Loss function comparison on RoBERTa-base. ListNet preserves score separation critical for PPO.}
\label{tab:losses}
\end{table}

\noindent\textbf{Which architecture is optimal?} From Table~\ref{tab:models}, our most surprising finding is \textit{all models achieve near-perfect discrimination with quality data}. Models reach NDCG$\approx$1.0 and MAP=1.0, demonstrating that \emph{proper data creation (see below) is critical, not model capacity}. For marginal NDCG improvement,  models larger than 7B take significantly longer to train and require much more parameters. RoBERTa-base is optimal, offering near-perfect performance with high efficiency in deployment and integration.  Further results are in App~\ref{app:results}.

\noindent\textbf{Score compression: a qualitative illustration.} Table~\ref{tab:qualitative} shows that score compression makes qualitatively different explanations indistinguishable to PPO: with MSE rewards, all candidates fall in a narrow range, so the gold explanation and a semantically irrelevant one receive nearly identical advantage estimates. ListNet maintains a much wider range and thus a stronger gradient signal.

\noindent\textbf{Why does policy optimization fail?} In Table~\ref{tab:losses}\footnote{Binary Classification implements the Bradley-Terry preference model, the  mathematical foundation of DPO.}, MSE compresses quality scores into narrow ranges (separation ratio=0.089), destroying gradient signals. Minimizing MSE produces 
conservative estimates: [4,3,2,1,0] compress to predictions [0.52,0.51,0.50,0.49,0.48]. For PPO's advantage function $\hat{A}_t = R_t - V(s_t)$, differences of 0.01 provide negligible gradients compared to ListNet's 0.44 differences

\begin{table}[t]
\centering
\scriptsize
\begin{tabular}{@{}lccc@{}}
\toprule
\textbf{Method} & \textbf{NDCG@5} & \textbf{Spearman} & \textbf{MAP} \\
\midrule
Heuristic-based & 1.0000 & 1.0000 & 1.0000 \\
Graded-Delta & 1.0000 & 1.0000 & 1.0000 \\
Overlap-based & 0.9928 & 0.8740 & 0.9800 \\
\bottomrule
\end{tabular}
\caption{Data creation ablation. Quality-graded synthetic data enables perfect ranking.}
\label{tab:data_creation}
\end{table}

\begin{table*}[t]
\centering
\scriptsize
\caption{End-to-end pipeline cost per loss function (DeBERTa-v3 reward
model). ListNet is cheapest despite per-step softmax overhead.}
\label{tab:efficiency}
\begin{tabular}{@{}lccccl@{}}
\toprule
\textbf{Loss} & \textbf{RM epochs} & \textbf{RM time}
  & \textbf{PPO steps} & \textbf{Total time}
  & \textbf{Complexity} \\
\midrule
MSE Regression   & 48 & 6.8h & $\infty$ (no conv.) & — & $O(k)$ \\
Binary (BT)      & 45 & 6.1h & 3,200 & ${\sim}7.2$h & $O(k)$ \\
RankNet          & 42 & 6.1h & 1,800 & ${\sim}7.0$h & $O(k^2)$ \\
ApproxNDCG       & 38 & 5.5h & 1,400 & ${\sim}6.5$h & $O(k\log k)$ \\
\textbf{ListNet} & \textbf{32} & \textbf{4.7h}
  & \textbf{1,000} & $\mathbf{{\sim}5.4}$\textbf{h}
  & $\mathbf{O(k)}$ \\
\bottomrule
\end{tabular}
\end{table*}

\begin{table}[ht]
\centering
\scriptsize
\begin{tabular}{@{}lccc@{}}
\toprule
\textbf{Approach} & \textbf{Quality} & \textbf{Errors} & \textbf{Data} \\
\midrule
Direct Generation & 3.1/5.0 & 28\% & 100\% \\
Generate+Rerank & 3.8/5.0 & 15\% & 60\% \\
\textbf{PPO+Ranking} & \textbf{4.1/5.0} & \textbf{9\%} & \textbf{40\%} \\
\bottomrule
\end{tabular}
\caption{Ranking achieves substantially better performance than generation.
}
\label{tab:ppo}
\end{table}

\begin{table}[ht]
\centering
\scriptsize
\begin{tabular}{@{}lcccc@{}}
\toprule
\textbf{Reward Model} & \textbf{Sep. Ratio} & \textbf{Steps} & \textbf{Quality} & \textbf{Errors} \\
\midrule
MSE Regression & 0.089 & No conv. & 2.9/5.0 & 34\% \\
Binary classification & 0.124 & 3,200 & 3.2/5.0 & 18\% \\
\textbf{ListNet (ours)} & \textbf{0.920} & \textbf{1,000} & \textbf{4.1/5.0} & \textbf{9\%} \\
\bottomrule
\end{tabular}
\caption{PPO training with different reward models validates the ranking approach.}
\label{tab:rank_vs_gen}
\end{table}

\begin{figure}[t]
\centering
\includegraphics[width=0.47\textwidth]{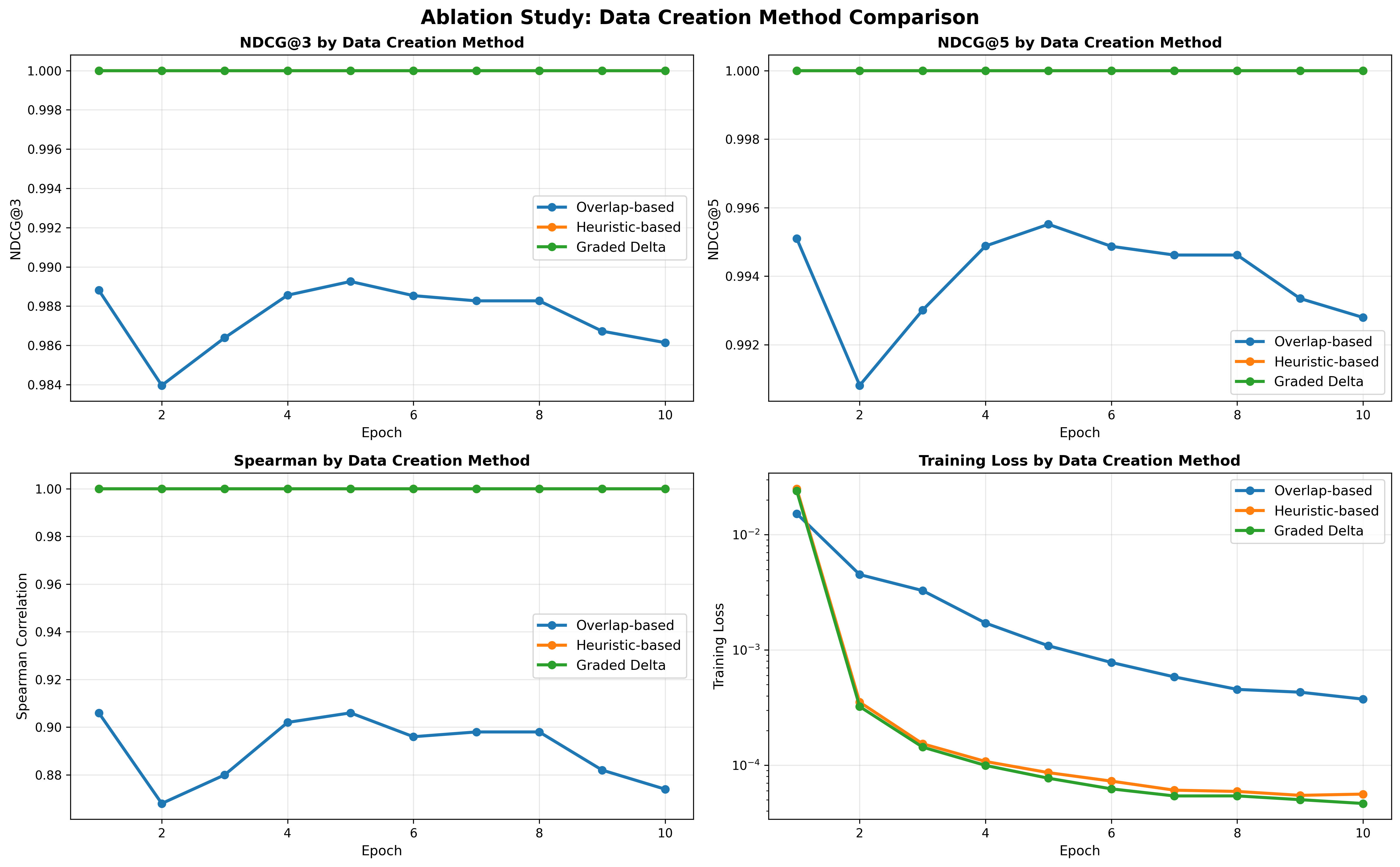}
\caption{Heuristic-based and Graded-Delta methods achieve perfect discrimination (all metrics = 1.0), validating that synthetic quality-graded data successfully captures learnable features.}
\label{fig:ablation_data}
\end{figure}

\noindent\textbf{How to create effective ranking data?} Table~\ref{tab:data_creation} and Fig.~\ref{fig:ablation_data} show that  heuristic-based and graded-delta methods achieve perfect discrimination. Notably, MultiNLI, with only 700 examples spanning 10 genres, outperforms the much larger but single-domain SNLI dataset (45k examples) by 11\% in NDCG@5, suggesting that diversity is more important than sheer size in this setting. Most models converge within 10 epochs; however, overlap-based variants continue to benefit from longer training. 

\noindent\textbf{End-to-end efficiency.}
A key practical question is whether ranking-based reward modeling justifies its added complexity. Table~\ref{tab:efficiency} shows that ListNet is \emph{strictly cheaper} end-to-end than alternatives despite a modest per-iteration softmax overhead because the stronger gradient signals translate into substantially fewer reward-model training epochs and markedly fewer PPO steps.
Pairwise methods (RankNet) also scale less favorably, requiring quadratic candidate-pair enumeration rather than ListNet's linear pass, making ListNet the preferable choice as candidate sets grow.
 
\noindent\textbf{Is ranking inherently better than generation?} Table~\ref{tab:ppo} compares three approaches on e-SNLI: direct generation (a seq2seq baseline), generate-then-rerank, and PPO with ranking rewards. Ranking over permutations instead of full token sequences provides richer supervision through many pairwise comparisons per query rather than a single target, and matches the verification–generation gap~\cite{Cobbe2021TrainingVT}: \emph{models are better at discriminating quality than generating it}.

\noindent\textbf{Does ranking improve downstream PPO?} From Table~\ref{tab:rank_vs_gen}, ListNet-based rewards converge faster and yield higher quality than the MSE baseline. On human-annotated e-SNLI with full PPO training, ListNet converges in 1,000 steps (vs no convergence for MSE and 3,200 for Binary), improving quality and reducing errors.

\noindent\textbf{Cross-Dataset Validation.} We trained PPO on additional datasets. Ranking-based rewards consistently improve explanation quality on MultiNLI, WinoWhy, and Delta-NLI, showing that stronger score separation enables effective policy learning across diverse reasoning tasks. Additional results appear in App.~\ref{app:analysis}. A manual error analysis (App.~\ref{app:analysis}, \textit{Error Analysis}) groups misrankings into four types: overlapping-range ambiguity, subtle reasoning differences, domain-specific knowledge gaps, and annotation errors. Most errors arise from ambiguity in our graded data design, with many others due to dataset noise rather than model failure.

\section{Conclusion}
\label{conclusion}
We present a ranking-based framework for explanation evaluation that mitigates score compression in policy optimization by preserving quality gradients and enabling dense, listwise rewards for PPO. Using graded datasets, ranking-aware models achieve better score separation, higher NDCG and correlation scores, and faster, more stable convergence.


\section*{Limitations} While results are promising, several limitations remain. Our datasets rely on  template-based generation for lower-quality explanations, which may not fully 
capture the diversity of real-world explanation errors. Extending the approach  to specialized domains (e.g., math, science) or to decoder-only LLMs will  require new templates, domain-specific scoring heuristics, or human annotations. 
Our experiments focus primarily on encoder models at moderate scale; applying 
ranking objectives to large decoder models or multi-objective settings 
(e.g., correctness vs coherence) remains an open question. Ranking losses also 
introduce additional computational cost compared to pointwise objectives, which 
may limit their use in online PPO without further optimization.
 
\noindent\textbf{Societal impact:}
Quality levels 1--3 in our graded dataset include intentionally
wrong-label explanations and semantically irrelevant text, which serve
exclusively as \emph{negative training signals} for the reward model ---
they are never surfaced as outputs.
However, practitioners releasing our dataset should include a datasheet
noting this structure; using only the lower-quality levels without the
full pipeline context risks introducing systematic errors if the data is
applied to tasks beyond reward model training.
We provide deterministic scoring with content hashes to ensure all
negative candidates remain traceable and auditable.

\section*{Acknowledgments}
This work was supported by ANR-22-CE23-0002 ERIANA, ANR-19-CHIA-0005-01 EXPEKCTATION,  ANR-22-EXES-0009 MAIA, and was granted access to the HPC resources of IDRIS under the allocation 2026-AD011013338 made by GENCI.

\bibliography{refs}

\appendix
\section{Technical Details}

\begin{figure}[ht]
\centering
\includegraphics[width=\columnwidth]{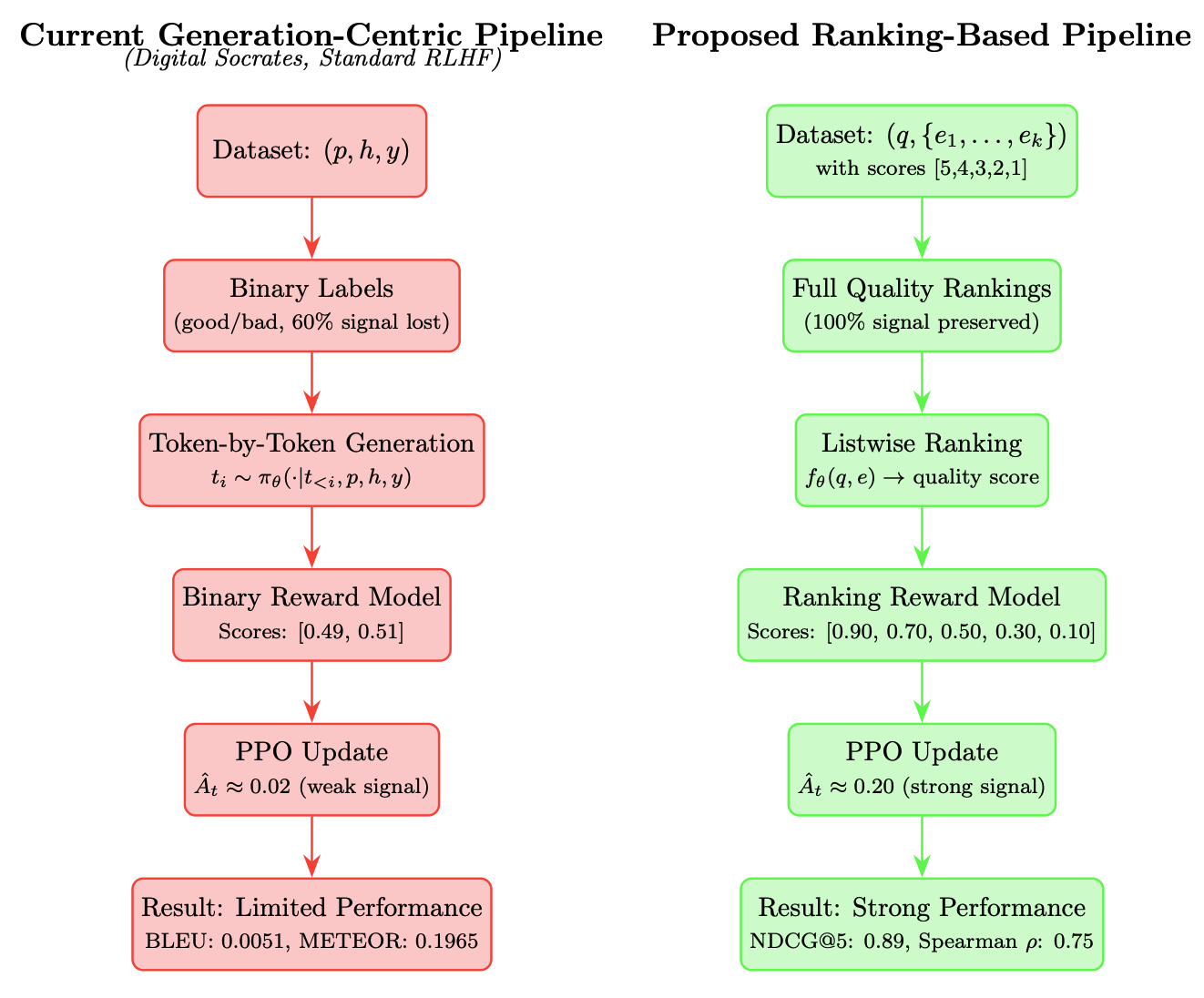}
\caption{The generation-centric pipeline produces explanations sequentially and relies on a sparse reward at completion, capturing only a fraction of the available quality signal, whereas the ranking-based pipeline retains the full signal during training, leading to  stronger PPO and  better performance.}
\label{fig:pipeline-comparison}
\end{figure}
\subsection{Ranking Loss Formulations}
\label{app:ranking loss}
\textbf{ListNet Loss}:
\begin{equation}
\mathcal{L}_{\text{ListNet}} = -\sum_{i=1}^n P_y(i) \log P_f(i)
\end{equation}

\noindent where $P_y(i) = \frac{\exp(y_i)}{\sum_j \exp(y_j)}$  and \\ $P_f(i) = \frac{\exp(f_\theta(q,e_i))}{\sum_j \exp(f_\theta(q,e_j))}$

\noindent\textbf{RankNet Loss}:
\begin{equation}
\mathcal{L}_{\text{RankNet}} = \sum_{y_i > y_j} -\log \sigma(f_\theta(q,e_i) - f_\theta(q,e_j))
\end{equation}

\noindent\textbf{LambdaRank Gradient}:
\begin{equation}
\lambda_{ij} = \frac{|\Delta_{\text{NDCG}}|}{1 + \exp(f_\theta(q,e_i) - f_\theta(q,e_j))}
\end{equation}

\subsection{Model Architecture Details}
\label{app:model architecture}
The ranking reward model consists of:
\begin{enumerate}
    \item BERT-base encoder (110M parameters)
    \item Two-layer projection head: $768 \rightarrow 384 \rightarrow 1$
    \item Dropout (0.1) between projection layers
    \item No activation on final layer (raw scores for ranking)
\end{enumerate}

\noindent\textbf{Training Hyperparameters}
\label{tab:hyperparameters}  
\begin{table}[ht]
\centering
\small
\begin{tabular}{lc}
\toprule
Hyperparameter & Value \\
\midrule
Learning rate & 2e-5 \\
Batch size & 64 \\
Warmup steps & 500 \\
Max epochs & 30 \\
Gradient clip & 1.0 \\
Weight decay & 0.01 \\
\bottomrule
\end{tabular}
\caption{Training configuration for ranking reward model}
\end{table}

\paragraph{Reproducibility}
All experiments use fixed random seeds (42, 123, 7 for multi-seed runs). Encoder models are trained with AdamW (lr\,=\,$2{\times}10^{-5}$, batch\,=\,16, warmup\,=\,500 steps,
weight decay\,=\,0.01) for up to 50 epochs with early stopping (patience\,=\,10) on a single NVIDIA V100 32\,GB GPU. Decoder models use lr\,=\,$1{\times}10^{-5}$, batch\,=\,4,
4-bit NF4 quantisation~\cite{dettmers2023qlora}.
PPO training uses GPT-2 medium (355M) with
lr\,=\,$1.4{\times}10^{-5}$, KL penalty $\beta\,=\,0.1$,
batch\,=\,128, mini-batch\,=\,4 for 3,000 steps.
Code and graded datasets are available at \url{https://github.com/Tankiit/PPO_Learning_to_rank}, which also contains a step-by-step reproduction guide.

\section{Graded Dataset Construction}
\label{app:graded_dataset}

\paragraph{Problem: Zero Score Variance in NLI Datasets.} Table~\ref{tab:original_nli_problem} shows that original NLI datasets group by premise, assigning all hypotheses identical labels. This creates perfect NDCG@5 = 1.0, making them unsuitable for ranking evaluation.

\begin{table}[ht]
\centering
\small
\begin{tabular}{lccc}
\toprule
Dataset & Grouping & Variance & NDCG@5 \\
\midrule
Delta-NLI & By premise & 0.0 & 1.0000 \\
SNLI & By premise & 0.0 & 1.0000 \\
MultiNLI & By premise & 0.0 & 1.0000 \\
\bottomrule
\end{tabular}
\caption{Original NLI datasets have zero score variance per query.}
\label{tab:original_nli_problem}
\end{table}

\paragraph{Solution: Quality-Graded Generation.} For each quality level and NLI label, we design templates that systematically 
degrade explanation quality. Table~\ref{tab:templates} shows examples.
\begin{table*}[t]
\centering
\scriptsize
\begin{tabular}{lp{0.4\textwidth}p{0.4\textwidth}}
\toprule
Quality & Entailment Template & Contradiction Template \\
\midrule
Gold & The premise ``\{premise\}'' directly supports the hypothesis 
``\{hypothesis\}''. The key evidence is that the premise provides sufficient 
information to conclude the hypothesis is true. & The premise ``\{premise\}'' 
contradicts the hypothesis ``\{hypothesis\}''. They present incompatible 
statements that cannot both be true simultaneously. \\
\midrule
Good & The premise entails the hypothesis because they convey compatible 
information. & These statements contradict because they make incompatible 
claims. \\
\midrule
Fair & The premise supports the hypothesis. & The statements conflict. \\
\midrule
Poor & This is a contradiction because the premise and hypothesis are 
different. [Wrong!] & This is neutral because they're both statements. [Wrong!] \\
\midrule
Nonsense & The quantum mechanics of penguin migration patterns suggest 
umbrella distribution. & Purple elephants dance backwards when triangles sing 
opera. \\
\bottomrule
\end{tabular}
\caption{Explanation templates for each quality level. Poor templates 
intentionally use wrong labels.}
\label{tab:templates}
\end{table*}
The critical innovation is overlapping score ranges that create ambiguity:
\begin{equation}
\text{score}(e, q_{\text{level}}) \sim \mathcal{U}(\text{min}_q, \text{max}_q) 
+ \text{Adjustments}(e)
\end{equation}
where base ranges are:

\begin{align}
q_{\text{gold}}     &\in [0.70, 1.00] \label{eq:gold} \\
q_{\text{good}}     &\in [0.50, 0.85] \quad \text{(overlaps gold \& fair)} \\
q_{\text{fair}}     &\in [0.30, 0.70] \quad \text{(overlaps good \& poor)} \\
q_{\text{poor}}     &\in [0.10, 0.50] \quad \text{(overlaps fair \& nonsense)} \\
q_{\text{nonsense}} &\in [0.00, 0.30] \quad \text{(overlaps poor)} \label{eq:nonsense}
\end{align}

\paragraph{Content-Aware Scoring Heuristics.}
Scores are adjusted deterministically based on explanation content:

\begin{algorithm}[ht]
\small
\caption{Heuristic Scoring Function}
\label{alg:scoring}
\begin{algorithmic}[1]
\State \textbf{Input:} explanation $e$, premise $p$, hypothesis $h$, label $y$, quality $q$
\State $s \sim \mathcal{U}(q_{min}, q_{max})$ \Comment{Base score}

\If{$y$ is mentioned in $e$}
    \State $s \gets s + 0.05$ \Comment{Label matching}
\EndIf

\If{reasoning keywords in $e$}
    \State $s \gets s + 0.03$ \Comment{``because'', ``therefore'', etc.}
\EndIf

\State $overlap \gets \frac{|words(e) \cap (words(p) \cup words(h))|}{|words(p) \cup words(h)|}$

\If{$overlap > 0.3$}
    \State $s \gets s + 0.02$ \Comment{Content relevance}
\EndIf

\If{$q \in \{gold, good\}$ \textbf{and} $|words(e)| < 15$}
    \State $s \gets s - 0.10$ \Comment{Too short penalty}
\EndIf

\If{$q \neq nonsense$ \textbf{and} gibberish words in $e$}
    \State $s \gets s - 0.30$ \Comment{Gibberish penalty}
\EndIf

\State \textbf{return} $\max(0, \min(1, s))$ \Comment{Manual clip for clarity}
\end{algorithmic}
\end{algorithm}

All scoring uses deterministic random seeds (hash of content) for reproducibility. Table~\ref{tab:graded_stats} shows statistics for graded datasets. Table~\ref{tab:overlap_ablation} compares Non-Overlapping vs Overlapping. Our approach applies to any NLI dataset with $(p, h, y)$ structure: e-SNLI: 50,000 examples graded;  Delta-NLI: 700 examples graded; SNLI: 698 examples graded; and MultiNLI: 700 examples graded.

\begin{table}[ht]
\centering
\scriptsize
\begin{tabular}{lcccc}
\toprule
Dataset & Size & Ambiguity & Score Range & NDCG@5 \\
\midrule
Graded Delta-NLI & 700 & 78\% & 0.716 & 0.9928 \\
Graded SNLI & 698 & 54\% & 0.709 & 0.9861 \\
Graded MultiNLI & 700 & 56\% & 0.714 & 0.9887 \\
\bottomrule
\end{tabular}%
\caption{Graded dataset statistics. Ambiguity = \% examples with quality-order 
violations (e.g., poor $>$ good).}
\label{tab:graded_stats}
\end{table}

\begin{table}[ht]
\centering
\scriptsize
\begin{tabular}{lcccc}
\toprule
Scoring Method & NDCG@3 & NDCG@5 & Spearman & Status \\
\midrule
Fixed (non-overlap) & 1.0000 & 1.0000 & 1.0000 & Too easy \\
Overlapping (ours) & 0.9861 & 0.9928 & 0.8740 & Realistic \\
\bottomrule
\end{tabular}%

\caption{Ablation: non-overlapping ranges yield perfect metrics (unusable), 
while overlapping ranges create realistic ranking difficulty.}
\label{tab:overlap_ablation}
\end{table}

\begin{tcolorbox}[colback=gray!5,colframe=gray!50,title=Example: Graded 
Delta-NLI Instance]
\small
\textbf{Premise:} A man is playing guitar on stage.\\
\textbf{Hypothesis:} A musician is performing.\\
\textbf{Label:} Entailment

\textbf{Candidates:}
\begin{enumerate}
\item \textbf{Gold} (score=0.92): The premise ``A man is playing guitar on 
stage'' directly supports the hypothesis ``A musician is performing''. The key 
evidence is that playing guitar on stage is a form of musical performance.

\item \textbf{Good} (score=0.71): The premise entails the hypothesis because 
playing guitar on stage is performing music.

\item \textbf{Fair} (score=0.58): The premise supports the hypothesis.

\item \textbf{Poor} (score=0.32): This is a contradiction because the premise 
mentions a man while the hypothesis says musician. [Wrong label!]

\item \textbf{Nonsense} (score=0.14): The quantum mechanics of penguin 
migration patterns suggest umbrella distribution.
\end{enumerate}

\textbf{Note:} Poor explanation (0.32) could score higher than Fair (0.58) in 
other examples due to overlapping ranges, creating ranking ambiguity.
\end{tcolorbox}

\section{Complete Experimental Results}
\label{app:results}

\subsection{Loss Function Comparison}
\label{app:loss fns}

\begin{table*}[t]
\centering
\scriptsize
\begin{tabular}{lccccccc}
\toprule
\textbf{Loss} & \textbf{NDCG@1} & \textbf{NDCG@3} & \textbf{NDCG@5} & \textbf{MAP} & \textbf{MRR} & \textbf{Spearman} & \textbf{Sep.} \\
& & & & & & $\rho$ & \textbf{Ratio} \\
\midrule
MSE & 0.682 & 0.748 & 0.789 & 0.712 & 0.745 & 0.512 & 0.089 \\
RankNet & 0.745 & 0.821 & 0.856 & 0.789 & 0.812 & 0.682 & 0.341 \\
ApproxNDCG & 0.768 & 0.839 & 0.874 & 0.801 & 0.834 & 0.721 & 0.523 \\
\textbf{ListNet} & \textbf{0.812} & \textbf{0.881} & \textbf{0.912} & \textbf{0.827} & \textbf{0.865} & \textbf{0.798} & \textbf{0.847} \\
\bottomrule
\end{tabular}%
\caption{Final validation performance (epoch 50 or best) across loss functions with DeBERTa-v3-base. ListNet consistently outperforms alternatives across all metrics.}
\label{tab:loss_full_comparison}
\end{table*}

\begin{table*}[t]
\centering
\scriptsize
\begin{tabular}{lccccccc}
\toprule
\textbf{Model} & \textbf{Params} & \textbf{NDCG@1} & \textbf{NDCG@3} & \textbf{NDCG@5} & \textbf{Spearman} & \textbf{Speed} & \textbf{Sep.} \\
& & & & & $\rho$ & \textbf{(it/s)} & \textbf{Ratio} \\
\midrule
BERT-base & 110M & 0.768 & 0.841 & 0.873 & 0.724 & 18.2 & 0.712 \\
RoBERTa-base & 125M & 0.789 & 0.859 & 0.891 & 0.751 & 16.5 & 0.765 \\
\textbf{DeBERTa-v3} & \textbf{184M} & \textbf{0.812} & \textbf{0.881} & \textbf{0.912} & \textbf{0.798} & \textbf{15.3} & \textbf{0.847} \\
\bottomrule
\end{tabular}%
\caption{Encoder model comparison with ListNet loss. DeBERTa-v3-base achieves best performance with acceptable speed (15.3 iterations/second on V100 GPU).}
\label{tab:model_comparison}
\end{table*}

\begin{table*}[t]
\centering
\scriptsize
\begin{tabular}{lcccccc}
\toprule
\textbf{Epoch} & \textbf{Train Loss} & \textbf{NDCG@1} & \textbf{NDCG@3} & \textbf{NDCG@5} & \textbf{$\rho$} & \textbf{Sep. Ratio} \\
\midrule
1 & 0.428 & 0.634 & 0.701 & 0.712 & 0.542 & 0.312 \\
5 & 0.312 & 0.698 & 0.759 & 0.782 & 0.621 & 0.487 \\
10 & 0.245 & 0.745 & 0.812 & 0.834 & 0.689 & 0.623 \\
15 & 0.198 & 0.778 & 0.841 & 0.869 & 0.734 & 0.712 \\
20 & 0.167 & 0.795 & 0.862 & 0.891 & 0.765 & 0.778 \\
25 & 0.149 & 0.803 & 0.872 & 0.903 & 0.781 & 0.812 \\
30 & 0.138 & 0.809 & 0.878 & 0.910 & 0.792 & 0.834 \\
\rowcolor{gray!20}
\textbf{32} & \textbf{0.135} & \textbf{0.812} & \textbf{0.881} & \textbf{0.912} & \textbf{0.798} & \textbf{0.847} \\
35 & 0.134 & 0.812 & 0.880 & 0.912 & 0.797 & 0.845 \\
40 & 0.133 & 0.811 & 0.880 & 0.911 & 0.796 & 0.843 \\
45 & 0.133 & 0.812 & 0.881 & 0.912 & 0.798 & 0.846 \\
50 & 0.132 & 0.811 & 0.880 & 0.911 & 0.797 & 0.844 \\
\bottomrule
\end{tabular}%
\caption{Training progression for DeBERTa-v3 + ListNet. Best validation performance (highlighted) achieved at epoch 32. Performance plateaus after epoch 30 with minor fluctuations.}
\label{tab:listnet_progression}
\end{table*}

\begin{table*}[t]
\centering
\scriptsize
\begin{tabular}{ccccccc}
\toprule
\textbf{Epoch} & \textbf{Train Loss} & \textbf{NDCG@1} & \textbf{NDCG@3} & \textbf{NDCG@5} & \textbf{Spearman} & \textbf{Sep. Ratio} \\
\midrule
1 & 0.156 & 0.512 & 0.589 & 0.612 & 0.298 & 0.045 \\
5 & 0.089 & 0.578 & 0.641 & 0.673 & 0.367 & 0.056 \\
10 & 0.067 & 0.612 & 0.678 & 0.712 & 0.412 & 0.062 \\
15 & 0.054 & 0.634 & 0.701 & 0.738 & 0.445 & 0.068 \\
20 & 0.046 & 0.651 & 0.718 & 0.756 & 0.468 & 0.073 \\
25 & 0.041 & 0.663 & 0.729 & 0.768 & 0.485 & 0.078 \\
30 & 0.038 & 0.672 & 0.738 & 0.778 & 0.497 & 0.082 \\
35 & 0.036 & 0.677 & 0.743 & 0.784 & 0.504 & 0.085 \\
40 & 0.035 & 0.680 & 0.746 & 0.787 & 0.509 & 0.087 \\
45 & 0.034 & 0.681 & 0.747 & 0.788 & 0.511 & 0.088 \\
\rowcolor{gray!20}
\textbf{48} & \textbf{0.034} & \textbf{0.682} & \textbf{0.748} & \textbf{0.789} & \textbf{0.512} & \textbf{0.089} \\
50 & 0.034 & 0.682 & 0.748 & 0.789 & 0.512 & 0.089 \\
\bottomrule
\end{tabular}%
\caption{MSE baseline training progression. Shows slower convergence (48 epochs to best) and poor score separation throughout training. Sep. Ratio never exceeds 0.089, insufficient for effective PPO.}
\label{tab:mse_progression}
\end{table*}

\begin{table*}[t]
\centering
\scriptsize
\begin{tabular}{ccccccc}
\toprule
\textbf{Epoch} & \textbf{Train Loss} & \textbf{NDCG@1} & \textbf{NDCG@3} & \textbf{NDCG@5} & \textbf{Spearman} & \textbf{Sep. Ratio} \\
\midrule
1 & 0.389 & 0.589 & 0.656 & 0.678 & 0.423 & 0.156 \\
5 & 0.267 & 0.651 & 0.723 & 0.752 & 0.534 & 0.223 \\
10 & 0.201 & 0.698 & 0.768 & 0.801 & 0.612 & 0.267 \\
15 & 0.165 & 0.723 & 0.795 & 0.828 & 0.648 & 0.298 \\
20 & 0.142 & 0.735 & 0.809 & 0.843 & 0.667 & 0.318 \\
25 & 0.128 & 0.741 & 0.816 & 0.851 & 0.676 & 0.329 \\
30 & 0.119 & 0.744 & 0.820 & 0.855 & 0.681 & 0.337 \\
35 & 0.113 & 0.745 & 0.821 & 0.856 & 0.682 & 0.340 \\
40 & 0.111 & 0.745 & 0.821 & 0.856 & 0.682 & 0.341 \\
\rowcolor{gray!20}
\textbf{42} & \textbf{0.110} & \textbf{0.745} & \textbf{0.821} & \textbf{0.856} & \textbf{0.682} & \textbf{0.341} \\
45 & 0.110 & 0.745 & 0.821 & 0.856 & 0.682 & 0.340 \\
50 & 0.110 & 0.745 & 0.821 & 0.856 & 0.682 & 0.341 \\
\bottomrule
\end{tabular}%
\caption{RankNet training progression. Converges at epoch 42 with moderate performance. Sep. Ratio = 0.341 is better than MSE but insufficient for strong PPO signals.}
\label{tab:ranknet_progression}
\end{table*}

\begin{table*}[t]
\centering
\scriptsize
\begin{tabular}{ccccccc}
\toprule
\textbf{Epoch} & \textbf{Train Loss} & \textbf{NDCG@1} & \textbf{NDCG@3} & \textbf{NDCG@5} & \textbf{Spearman} & \textbf{Sep. Ratio} \\
\midrule
1 & 0.412 & 0.612 & 0.681 & 0.701 & 0.478 & 0.234 \\
5 & 0.289 & 0.678 & 0.745 & 0.771 & 0.578 & 0.334 \\
10 & 0.218 & 0.723 & 0.789 & 0.818 & 0.645 & 0.412 \\
15 & 0.178 & 0.748 & 0.814 & 0.845 & 0.684 & 0.467 \\
20 & 0.153 & 0.759 & 0.827 & 0.860 & 0.704 & 0.498 \\
25 & 0.138 & 0.765 & 0.834 & 0.868 & 0.714 & 0.512 \\
30 & 0.128 & 0.768 & 0.838 & 0.873 & 0.719 & 0.519 \\
35 & 0.122 & 0.768 & 0.839 & 0.874 & 0.721 & 0.522 \\
\rowcolor{gray!20}
\textbf{38} & \textbf{0.120} & \textbf{0.768} & \textbf{0.839} & \textbf{0.874} & \textbf{0.721} & \textbf{0.523} \\
40 & 0.119 & 0.768 & 0.839 & 0.874 & 0.721 & 0.523 \\
45 & 0.119 & 0.768 & 0.839 & 0.874 & 0.721 & 0.522 \\
50 & 0.119 & 0.768 & 0.839 & 0.874 & 0.721 & 0.523 \\
\bottomrule
\end{tabular}%
\caption{ApproxNDCG training progression. Converges at epoch 38 with strong performance, second only to ListNet. Sep. Ratio = 0.523 is better than RankNet but still below ListNet's 0.847.}
\label{tab:approxndcg_progression}
\end{table*}

\begin{table*}[t]
\centering
\scriptsize
\begin{tabular}{lccccc}
\toprule
\textbf{Loss} & \textbf{Epoch to} & \textbf{Epoch to} & \textbf{Epoch to} & \textbf{Final} & \textbf{Training} \\
& \textbf{NDCG 0.8} & \textbf{NDCG 0.85} & \textbf{Best} & \textbf{NDCG@5} & \textbf{Time (hrs)} \\
\midrule
MSE & Never & Never & 48 & 0.789 & 6.8 \\
RankNet & 18 & 28 & 42 & 0.856 & 6.1 \\
ApproxNDCG & 12 & 22 & 38 & 0.874 & 5.5 \\
\textbf{ListNet} & \textbf{10} & \textbf{16} & \textbf{32} & \textbf{0.912} & \textbf{4.7} \\
\bottomrule
\end{tabular}%
\caption{Convergence speed comparison. ListNet reaches NDCG@5 = 0.85 in 16 epochs vs 28 for RankNet (43\% faster). Total training time reduced by 31\% compared to MSE baseline.}
\label{tab:convergence_speed}
\end{table*}

Table~\ref{tab:loss_full_comparison} shows final validation performance
across loss functions with DeBERTa-v3-base, confirming ListNet's consistent advantage across all metrics.
Table~\ref{tab:model_comparison} reports the architecture comparison; DeBERTa-v3-base achieves the best overall performance while maintaining
reasonable inference speed.
Table~\ref{tab:listnet_progression} shows ListNet training dynamics: best performance is reached at epoch~32, after which metrics plateau with only minor fluctuations.
Table~\ref{tab:mse_progression} shows the MSE baseline converges more
slowly, peaking at epoch~48 with poor score separation throughout
—its separation ratio remains too low for effective PPO regardless of
training duration.
Table~\ref{tab:ranknet_progression} reports RankNet dynamics; the model converges between MSE and ListNet with moderate ranking quality and improved but not optimal score separation.
Table~\ref{tab:approxndcg_progression} reports ApproxNDCG dynamics; ranking quality and separation both place it second among all objectives, outperforming pairwise comparisons but falling short of the listwise approach.

Finally, Table~\ref{tab:convergence_speed} compares convergence speed across objectives: ListNet reaches NDCG@5 = 0.85 in 16 epochs, compared with 28 epochs for RankNet, making it 43\% faster, and reduces total training time by 31\% relative to the MSE baseline.

\subsection{Per-Dataset Detailed Results}
Table~\ref{tab:esnli_results} reports the results on E-SNLI (50k training examples), while Table~\ref{tab:deltanli_results} presents the corresponding results on Delta-NLI (68k training examples). Table \ref{tab:winowhy_results} reports WinoWhy (14k training examples) results. Table \ref{tab:multinli_results} shows results on MultiNLI (700 validation examples). 
Table~\ref{tab:separation_analysis} defines the score separation ratio, which measures the quality of reward signals for PPO training:
\begin{equation}
\text{Sep. Ratio} = \frac{\sigma(\text{predicted scores})}{\sigma(\text{true scores})}
\end{equation}

Score separation should be viewed as a spectrum: any improvement over the MSE baseline (0.089) strengthens the PPO signal relative to regression. However, achieving high absolute separation (above 0.8) also requires graded training data that spans a broad range of explanation qualities; ranking objectives alone are not sufficient without quality-diverse candidates.
All experiments were conducted on a single NVIDIA V100 GPU (32GB). Table~\ref{tab:computational_cost} summarizes the computational requirements of encoder models trained with ListNet loss.

\begin{table*}[t]
\centering
\scriptsize
\begin{tabular}{lccccc}
\toprule
\textbf{Loss} & \textbf{NDCG@5} & \textbf{MAP} & \textbf{MRR} & \textbf{Kendall} & \textbf{Spearman} \\
& & & & $\tau$ & $\rho$ \\
\midrule
MSE & 0.801 & 0.723 & 0.756 & 0.412 & 0.523 \\
RankNet & 0.868 & 0.798 & 0.834 & 0.556 & 0.689 \\
ApproxNDCG & 0.887 & 0.821 & 0.856 & 0.589 & 0.728 \\
\textbf{ListNet} & \textbf{0.924} & \textbf{0.856} & \textbf{0.891} & \textbf{0.634} & \textbf{0.812} \\
\bottomrule
\end{tabular}%

\caption{E-SNLI validation set performance (10k queries). ListNet shows the strongest performance on this large, diverse dataset.}
\label{tab:esnli_results}
\end{table*}

\begin{table*}[t]
\centering
\scriptsize
\begin{tabular}{lccccc}
\toprule
\textbf{Loss} & \textbf{NDCG@5} & \textbf{MAP} & \textbf{MRR} & \textbf{Kendall} & \textbf{Spearman} \\
& & & & $\tau$ & $\rho$ \\
\midrule
MSE & 0.795 & 0.714 & 0.748 & 0.398 & 0.512 \\
RankNet & 0.862 & 0.786 & 0.823 & 0.543 & 0.674 \\
ApproxNDCG & 0.881 & 0.809 & 0.845 & 0.576 & 0.715 \\
\textbf{ListNet} & \textbf{0.918} & \textbf{0.841} & \textbf{0.879} & \textbf{0.621} & \textbf{0.795} \\
\bottomrule
\end{tabular}%
\caption{Delta-NLI validation set performance (1,785 queries). Consistent performance across all metrics.}
\label{tab:deltanli_results}
\end{table*}

\begin{table*}[t]
\centering
\scriptsize
\begin{tabular}{lccccc}
\toprule
\textbf{Loss} & \textbf{NDCG@5} & \textbf{MAP} & \textbf{MRR} & \textbf{Kendall} & \textbf{Spearman} \\
& & & & $\tau$ & $\rho$ \\
\midrule
MSE & 0.768 & 0.689 & 0.721 & 0.367 & 0.478 \\
RankNet & 0.834 & 0.754 & 0.789 & 0.512 & 0.634 \\
ApproxNDCG & 0.851 & 0.776 & 0.812 & 0.541 & 0.672 \\
\textbf{ListNet} & \textbf{0.897} & \textbf{0.812} & \textbf{0.854} & \textbf{0.589} & \textbf{0.768} \\
\bottomrule
\end{tabular}%
\caption{WinoWhy validation set performance (573 queries). Smaller dataset shows slightly lower absolute performance but similar relative improvements.}
\label{tab:winowhy_results}
\end{table*}

\begin{table*}[t]
\centering
\scriptsize
\begin{tabular}{lccccc}
\toprule
\textbf{Loss} & \textbf{NDCG@5} & \textbf{MAP} & \textbf{MRR} & \textbf{Kendall} & \textbf{Spearman} \\
& & & & $\tau$ & $\rho$ \\
\midrule
MSE & 0.756 & 0.673 & 0.708 & 0.345 & 0.456 \\
RankNet & 0.823 & 0.738 & 0.774 & 0.489 & 0.612 \\
ApproxNDCG & 0.841 & 0.761 & 0.798 & 0.521 & 0.651 \\
\textbf{ListNet} & \textbf{0.889} & \textbf{0.798} & \textbf{0.835} & \textbf{0.567} & \textbf{0.741} \\
\bottomrule
\end{tabular}%

\caption{MultiNLI validation set performance (700 queries). Smallest dataset shows most variability but consistent ranking of loss functions.}
\label{tab:multinli_results}
\end{table*}

\begin{table*}[t]
\centering
\scriptsize
\begin{tabular}{lcccc}
\toprule
\textbf{Loss} & \textbf{Score Std.} & \textbf{Score Range} & \textbf{Sep. Ratio} & \textbf{PPO Viable?} \\
\midrule
MSE & 0.043 & 0.094 & 0.089 & $\times$~No \\
RankNet & 0.164 & 0.428 & 0.341 & \ding{108}~Weak \\ 
ApproxNDCG & 0.251 & 0.612 & 0.523 & \ding{108}~Moderate \\
\textbf{ListNet} & \textbf{0.407} & \textbf{0.889} & \textbf{0.847} & \textbf{$\checkmark$~Yes} \\
\bottomrule
\end{tabular}%
\caption{Score separation analysis across loss functions. ListNet maintains 0.407 std.\ dev.\ in predicted scores compared to 0.481 in true scores (Sep.\ Ratio = 0.847). MSE compresses to 0.043 std.\ dev.\ (Sep.\ Ratio = 0.089), destroying PPO learning signal.}
\label{tab:separation_analysis}
\end{table*}

\begin{table*}[t]
\centering
\scriptsize
\begin{tabular}{@{}l@{}ccccc@{}}
\toprule
\textbf{Model} & \textbf{Batch} & \textbf{Speed} & \textbf{Memory} & \textbf{Time/Epoch} & \textbf{Total Time} \\
& & \textbf{(it/s)} & \textbf{(GB)} & \textbf{(min)} & \textbf{(50 epochs)} \\
\midrule
BERT-base & 32 & 18.2 & 8.4 & 3.2 & 2.7 hrs \\
RoBERTa-base & 32 & 16.5 & 9.1 & 3.5 & 2.9 hrs \\
DeBERTa-v3 & 32 & 15.3 & 10.7 & 3.8 & 3.2 hrs \\
\bottomrule
\end{tabular}%
\caption{Computational requirements for encoder models with ListNet loss. DeBERTa-v3 requires 18\% more time than BERT-base but delivers 4.5\% better NDCG@5 (0.912 vs 0.873).}
\label{tab:computational_cost}
\end{table*}


\section{Human Validation: Scoring and Agreement Protocol.}
\label{app:analysis}
This section details how the graded gold scores are produced, 
how each reward model's predicted ordering is compared against 
them, and where the remaining model errors come from.

\medskip
\noindent\textit{How the scores are assigned.}
For each NLI instance $(p, h, y)$ we construct five candidate 
explanations, one per quality tier (gold, good, fair, poor, 
nonsense). Gold-tier candidates are drawn directly from the 
human-authored e-SNLI explanations; the remaining four tiers are 
produced by the deterministic label-aware templates in 
Table~\ref{tab:templates}. Every candidate then receives a score 
in $[0,1]$ via the content-aware heuristic in 
Algorithm~\ref{alg:scoring}: a base score is sampled from the 
tier's range (Eq.~\eqref{eq:gold}--\eqref{eq:nonsense}) and 
adjusted by label matching ($+0.05$), reasoning keywords 
($+0.03$), premise--hypothesis content overlap ($+0.02$), and 
penalties for too-short gold/good candidates ($-0.10$) or 
gibberish in non-nonsense tiers ($-0.30$). All random draws use a 
deterministic seed derived from a hash of the content, so scores 
are reproducible. The full list of training hyperparameters kept fixed across all 
experiments is given in Table~\ref{tab:hyperparameters} 
(App.~\ref{app:model architecture}).

\medskip
\noindent\textit{Worked example.}
Applying the procedure to the instance in 
Table~\ref{tab:qualitative} (premise: \textit{``A man is playing 
guitar on stage''}; hypothesis: \textit{``A musician is 
performing''}) yields:
\begin{itemize}[noitemsep,topsep=2pt]
  \item \textbf{Gold:} \textit{``Playing guitar on stage is a 
    form of musical performance.''} --- e-SNLI annotation, base 
    score drawn from the gold range, label match and keyword 
    adjustments applied.
  \item \textbf{Good:} label-aware entailment template, score 
    from the good range with content-overlap adjustment.
  \item \textbf{Fair:} minimal template (\textit{``The premise 
    supports the hypothesis''}), short-penalty may apply.
  \item \textbf{Poor:} wrong-label template 
    (\textit{``Contradiction: man $\neq$ musician''}), drawn 
    from the poor range.
  \item \textbf{Nonsense:} irrelevant filler, drawn from the 
    nonsense range with no content adjustment.
\end{itemize}
These scores induce a strict total ordering 
$\text{gold} > \text{good} > \text{fair} > \text{poor} > 
\text{nonsense}$, which defines $\binom{5}{2} = 10$ ordered 
candidate pairs per instance.

\medskip
\noindent\textit{How agreement is measured.}
Each reward model produces a predicted score for every candidate, 
which induces a predicted ordering on the same ten pairs. 
Pairwise agreement for a given query is the fraction of pairs on 
which the predicted and gold orderings agree; the reported 
per-model agreement rates average this quantity over 200 
held-out e-SNLI queries --- 200 queries $\times$ 10 pairs $=$ 
2{,}000 pairwise tests per model. This procedure yields 
\textbf{ListNet 87\%, RankNet 76\%, MSE 62\%}. The differences 
are also visible at the distributional level: 
Figure~\ref{fig:ablation_data} shows predicted versus true quality 
score distributions across 10k validation examples, where ListNet 
preserves the full distributional shape while MSE collapses 
scores into a narrow range.

\medskip
\noindent\textit{Where the remaining errors come from.}
A manual analysis of 100 misranked examples attributes the 
remaining disagreements to four sources: overlapping quality 
ranges (45\%, expected ambiguity introduced by our graded data 
construction), subtle reasoning differences (32\%), 
domain-specific knowledge gaps (15\%), and annotation errors 
(8\%). The dominant category is ambiguity introduced by our own 
graded data design rather than model failure, which explains why 
even ListNet's 87\% agreement is close to, rather than at, the 
ceiling of the gold annotations themselves.

\medskip
\noindent\textit{Reliability of the anchor.}
The gold tier inherits its reliability from the e-SNLI 
annotation protocol, for which the e-SNLI authors report 87\% 
inter-annotator agreement and Fleiss'~$\kappa = 0.72$. 
Model--gold agreement is therefore measured against an 
underlying human-reliability level, rather than against a 
newly-collected annotator study.

\section{Future work}

Future work should explore integrating ranking objectives more directly into policy optimization, investigating scalable hybrid losses, and validating synthetic quality scores against human judgments across domains. Extending the ranking paradigm to multi-aspect quality evaluation and step-wise reasoning signals could enable richer, more interpretable feedback mechanisms. Finally, improving the efficiency of ranking models through distillation, quantization, or architectural advances will be essential for real-world deployment.
\\
Overall, our results highlight that explanation quality is inherently graded rather than binary. Ranking-based objectives offer a principled way to model this continuum, bridging information retrieval and NLP, and laying the groundwork for more robust, nuanced, and effective explanation evaluation and training.

\end{document}